\pdfoutput=1

\documentclass[nohyperref]{article}

\usepackage{xspace} 
\usepackage[dvipsnames]{xcolor}
\usepackage{enumitem}
\definecolor{myred}{RGB}{204, 0, 0}
\definecolor{mygreen}{RGB}{0, 99, 0}
\definecolor{myblue}{RGB}{0, 0, 99}
\definecolor{myyellow}{RGB}{204, 204, 0}

\usepackage{microtype}
\usepackage{graphicx}
\usepackage{subfigure}
\usepackage{booktabs} 

\usepackage{hyperref}

\usepackage[accepted]{icml2022}

\usepackage{amsmath}
\usepackage{amssymb}
\usepackage{mathtools}
\usepackage{amsthm}

\usepackage[capitalize,noabbrev]{cleveref}

\theoremstyle{plain}

\theoremstyle{definition}

\theoremstyle{remark}

\usepackage[textsize=tiny]{todonotes}

\icmltitlerunning{Revisiting Hotels-50K and Hotel-ID}

\begin{document}

\twocolumn[
\icmltitle{Revisiting Hotels-50K and Hotel-ID}







\begin{icmlauthorlist}
\icmlauthor{Aarash Feizi}{mcgill,mila}
\icmlauthor{Arantxa	Casanova}{pt,mila}
\icmlauthor{Adriana	Romero-Soriano}{mcgill,mila}
\icmlauthor{Reihaneh Rabbany}{mcgill,mila}

\end{icmlauthorlist}

\icmlaffiliation{mcgill}{McGill University, Montr\'eal, Canada}

\icmlaffiliation{mila}{Mila Institute, Montr\'eal, Canada}

\icmlaffiliation{pt}{\'Ecole Polytechnique de Montr\'eal, Montr\'eal, Canada}

\icmlcorrespondingauthor{Aarash Feizi}{aarash.feizi@mila.quebec}
\icmlkeywords{Machine Learning, ICML}

\vskip 0.3in
]



\printAffiliationsAndNotice{}  


\newcommand{\cut}[1]{}
\newcommand{\firstval}{\text{$\mathcal{D}^{val}_{SS}$}\xspace}
\newcommand{\secondval}{\text{$\mathcal{D}^{val}_{SU}$}\xspace}
\newcommand{\thirdval}{\text{$\mathcal{D}^{val}_{UU}$}\xspace}
\newcommand{\fourthval}{\text{$\mathcal{D}^{val}_{??}$}\xspace}

\newcommand{\firsttest}{\text{$\mathcal{D}_{SS}$}\xspace}
\newcommand{\secondtest}{\text{$\mathcal{D}_{SU}$}\xspace}
\newcommand{\thirdtest}{\text{$\mathcal{D}_{UU}$}\xspace}
\newcommand{\fourthtest}{\text{$\mathcal{D}_{??}$}\xspace}

\newcommand{\hotelsfifty}{Revisited Hotels-50K\xspace}
\newcommand{\hotelsid}{Revisited Hotel-ID\xspace}

\newcommand{\randauroc}{$\text{AUC}$\xspace}

\newcommand{\newauroc}{$\text{AUC}_{H}$\xspace}
\newcommand{\recallatone}{R@1\xspace}

\newcommand{\s}[2]{s(#1, #2)}

\begin{abstract}

In this paper, we propose revisited versions for two recent hotel recognition datasets: Hotels-50K and Hotel-ID. The revisited versions provide evaluation setups with different levels of difficulty to better align with the intended real-world application, i.e. countering human trafficking. Real-world scenarios involve hotels and locations that are not captured in the current data sets, therefore it is important to consider evaluation settings where classes are truly unseen. We test this setup using multiple state-of-the-art image retrieval models and show that as expected, the models’ performances decrease as the evaluation gets closer to the real-world unseen settings. The rankings of the best performing models also change across the different evaluation settings, which further motivates using the proposed revisited datasets. 

\end{abstract}

\vspace{-20pt}
\section{Introduction and Background}
\label{sec:intro}




Forced labor and human-trafficking (HT) are one of the biggest issues of our current society. The International Labour Organization estimates that this industry has an annual profit of \$99 billion worldwide \cite{international2017global}. With the advance of advertisement technology and easy-to-use and low-risk online platforms, traffickers have been exploiting such platforms for recruiting victims, e.g. by posting fake recruitment ads, or soliciting buyers, e.g. by advertising their victims on online escort websites. 

According to \cite{minor2015report}, the majority of victims of HT are advertised online; therefore, escort websites have become a reliable resource for anti-trafficking operations. However, due to the large volume of data on these websites, manually analyzing all the ads and identifying potential victims of HT is practically infeasible for law enforcement. Motivated by this, there has been a growing number of studies on developing data-driven algorithms for HT detection \cite{li2018detection,rabbany2018active,tong2017combating}.
\begin{figure}
    \centering
    \includegraphics[width=0.98\linewidth]{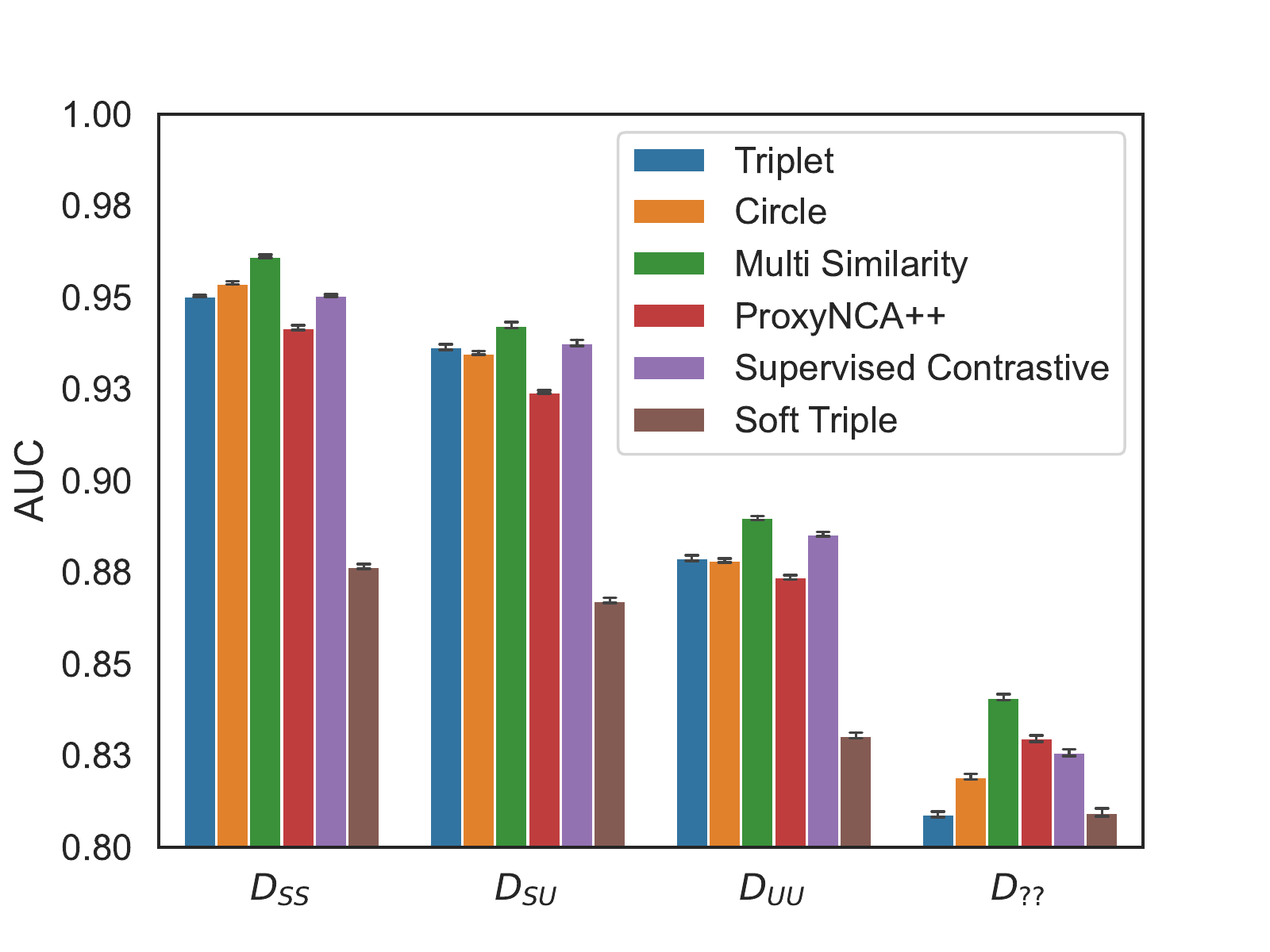}
    \vspace{-20pt}
    \caption{AUC in the proposed splits for \textbf{\hotelsfifty}, namely, \firsttest: \{branch: seen, chain: seen\}, \secondtest: \{branch: unseen, chain: seen\}, \thirdtest: \{branch: unseen, chain: unseen\} and  \fourthtest: \{branch: unknown, chain: unknown\}. All evaluated models experience a consistent drop for the unseen and unknown splits.}
    \label{fig:hotels50k-auc-barplot}\vspace{-15pt}
\end{figure}
Studies on data-driven HT detection rely on discovering certain information from individual or multiple ads that separate them from genuine ads and makes them suspicious of trafficking. Said information could be extracted from the text \cite{nagpal2015entity,li2018detection}, the images used in ads \cite{stylianou2019hotels}, or both \cite{tong2017combating}. 
Although there has been a growing amount of research on anti-HT, most of these studies focus on analyzing the textual data and overlook the images in their analysis \cite{li2018detection}, or fail to effectively incorporate the information form images, e.g. only looking at exact matches \cite{rabbany2018active} or having no performance gain \cite{tong2017combating}. 
%
%
Moreover, images can be considered as a group to draw connections between them. For example, if a set of images from different advertisements share a common background or a certain feature in their pictures, these advertisements can be flagged as ``suspicious and organized'' and could be further analyzed by an investigator.
Many image datasets have been proposed in previous literature; however, most of them are not suitable for the mentioned tasks of countering HT. In more detail, most images encountered in this domain contain masked people in different locations, e.g. hotel rooms, and are mostly similar to each other, whether they are taken in the same hotel or not. Hence, the difference between images usually lie in the subtle details of the objects that are present in most images, e.g. bed sheets, and datasets consisting of classes with evidently different objects, e.g. dogs, chairs, airplanes, do not suit this task. For this reason, Hotels-50K \cite{stylianou2019hotels} and Hotel-ID \cite{kamath20212021} were introduced for hotel classification for anti-HT, which contains images of hotel rooms from numerous hotel branches around the world. The datasets additionally provide the hotel chain, i.e., super-class, information for the hotel branches, i.e., classes. Refer to Appendix \ref{sec:background} for a more in-depth literature review.

Since Hotels-50K and Hotel-ID were collected and annotated for hotel classification, the hotel branches used for evaluation are a subset of hotel branches used for training. Nevertheless, this is far from the real-world situation where it is highly unlikely for all locations to be known during training the model. We propose a revised version of the Hotels-50K and Hotel-ID datasets, which consists of a single training set, with multiple evaluation splits with different levels of difficulty, e.g. as seen in Figure \ref{fig:hotels50k-auc-barplot}. The difficulty of each evaluation split is based on whether the classes and superclasses in a given split are present in the training set or not. This enables researchers to better evaluate baseline or state-of-the-art vision models and how well they will generalize to the real-world scenario and unseen data.

In order to evaluate our new data splits, we run extensive experiments with multiple recent image retrieval models. We show that the performance of all models decrease as the data splits become more realistic. Also, we show that the order of best performing methods differ between different levels of difficulty.

The main contributions of this paper are as follows:
\begin{itemize}[noitemsep,topsep=0pt,parsep=0pt,partopsep=0pt,leftmargin=*]
    \item We provide revisited versions of the Hotels-50K and Hotel-ID datasets for developing algorithms to counter human trafficking, namely \hotelsfifty and \hotelsid. These datasets resolve inconsistencies in the original data and also contain four levels of difficulty in order to replicate the real-world scenario and enable us to better evaluate different models.
    
    
    \item We repoert the performance of state-of-the-art image retrieval models on the proposed datasets, observing that their performance drops in line with the difficulty of the splits. Moreover, we find that the methods rank differently in each split, evidencing the relevance of our proposed splits, as the performance on the data split containing only seen classes during training does not predict the performance on the more challenging and realistic splits.
\end{itemize}
We provide the proposed revisited datasets and splits which control the task difficulty on GitHub, available at:  \url{https://github.com/aarashfeizi/revisited-hotels}, for the research community.




\vspace{-3pt}
\section{Datasets}
\label{sec:hotels}

Two of the most established large vision datasets in the current literature are ImageNet \cite{deng2009imagenet} and MS-COCO \cite{lin2014microsoft}. ImageNet is mostly used for pretraining vision models and evaluating new methods for image classification tasks, and MS-COCO is a generally used for object detection and image segmentation. However, due to the high diversity of objects used for creating these datasets, they have a high inter-class variance, i.e. different classes of objects have clear and evident visual differences among themselves. As such, models trained using them will tend to learn how to distinguish high-level visual features, e.g. object shapes, rather than detailed patterns on the same kind of objects. To train models for more fine-grained tasks, datasets with lower inter-class variance, such as CUB-2011-200 \cite{wah2011caltech} or Cars196 \cite{krause20133d}, are also proposed, in which the pattern or specific details on the birds or cars become important for classification. Hotels-50K \cite{stylianou2019hotels} and the Hotel-ID \cite{kamath20212021} are two of the most recent image datasets with a low inter-class variance, created specifically for applications in countering human trafficking.

\vspace{-5pt}
\subsection{Original Datasets} Hotels-50K \cite{stylianou2019hotels}, introduced in 2019, is one of the recent fine-grained datasets developed for hotel classification\cut{for the purpose of anti-HT}. For each hotel, this dataset also provides the hotel's chain information and also their geolocations. The images have been gathered from two sources: travel websites, such as Expedia, and TraffickCam, a crowdsourcing mobile application allowing its users to submit photos of their hotel room. In the original splits of the dataset, the training data consists of 50,000 hotel labels, which are further categorized into 91 hotel chains, if known. Albeit the train set is gathered from both sources, the test contains 4,718 classes gathered from only TraffickCam because of the more challenging nature, e.g. bad lighting or clutters, of the images. More recently, Hotel-ID \cite{kamath20212021} was introduced, which is a subset of the original Hotels-50K dataset and only contains images form TraffickCam. This smaller dataset contains 97,553 images from 7,769 hotels, which includes 86 of the known hotel chains from the Hotels-50K dataset.

\vspace{-5pt}
\subsection{Proposed Datasets}
\begin{figure}
    \centering
    \includegraphics[width=\linewidth]{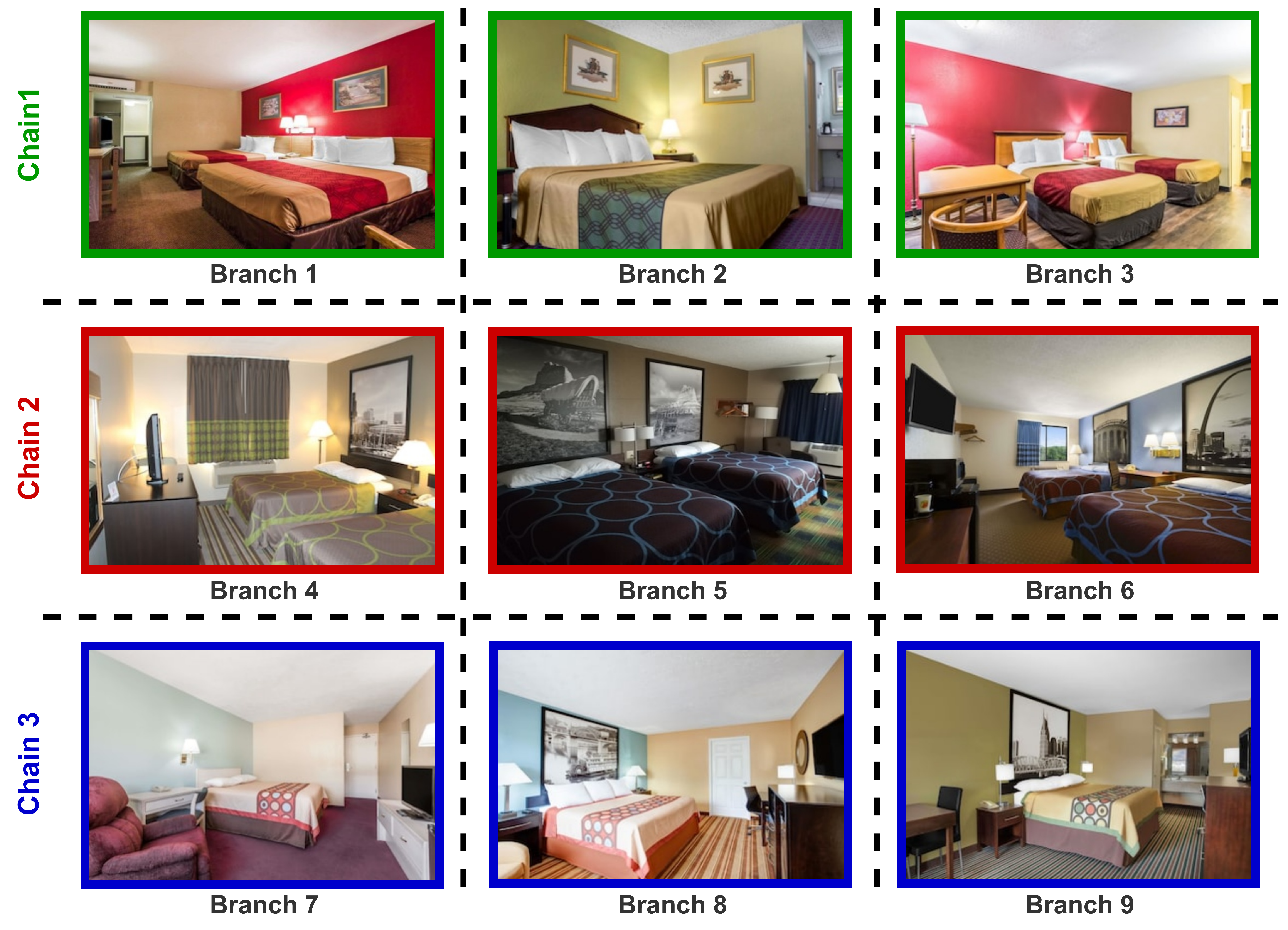}
         \vspace{-20pt}
 \caption{Sampled images from Hotels-50K, showcasing 9 \textbf{different} hotel branches (i.e. classes) and 3 hotel chains (i.e. super classes). Hotel branches from the same chain may have shared patterns (e.g. bedsheet styles) which motivates splitting the data by chains to have truly unseen classes. }
    \vspace{-19pt}
    \label{fig:hotel-same-chain}
\end{figure}

In previous studies \cite{stylianou2019hotels, xuan2020improved, kamath20212021}, Hotels-50K and Hotel-ID have mainly been used for hotel classification. This can be used for example to locate a victim based on an image. However, the data splits and categorization in these datasets are provided in such a way that it does not allow testing for unseen hotels or rooms, which would test the generalization of the models; mainly since: 1) All test classes are seen during training, meaning that the model will most likely train on all available patterns before being tested, which is not a real-world scenario. Even assuming we can collect a large dataset of images of all registered hotels, many of the locations traffickers use for taking pictures are not hotels, and in particular with the training of hospitality staff on identifying trafficking cases, there has been an increasing shift into alternative accommodations including Airbnbs \cite{thulemark2021sharing}. Therefore, it is not realistic to assume a model can be trained on all possible locations. 2) Many different hotels under the same hotel chain contain similar patterns/designs in their rooms. An anecdotal example is shown in Figure \ref{fig:hotel-same-chain}, which depicts 9 hotel branches from three hotel chains (rows). Note that different branches are considered different classes in this data set, and in this example, the bed sheets in each chain share the same pattern across branches. Hence, even having distinct hotel branches of the same hotel chain in both training and testing sets, causes the model to possibly learn complex visual patterns from hotel branches it was not trained on and defeats the purpose of testing the model on unseen patterns.    
Considering these concerns, we propose revised train and evaluations sets to satisfy the following primary criteria:
\begin{itemize}[noitemsep,topsep=0pt,parsep=0pt,partopsep=0pt,leftmargin=*]
    \item reserve a set of hotel \textit{branch} images only for evaluating,
    \item reserve a set of hotel \textit{chain} images only for evaluating.
\end{itemize}

These properties ensure that considering the training set, the evaluation sets recreate real-world scenarios for measuring a model's performance where we need to connect images from unseen/new locations. 
In addition to these two criteria, we also ensure that a minimum number of images exist per class in both training and evaluation sets, so the model has access to enough instances.

Since the main focus of the new splits is for the model to capture specific visual features from images and match them with other images, whether they are collected from TraffickCam or a travel website, it is not necessary for the test set to be comprised of only images from TraffickCam. Hence, we first merge the original training and testing sets to get a complete set of all the images from 50,000 hotels. Knowing one of the main image sources of this data is travel websites and is collected using website APIs, we checked for duplicate images under different classes and removed them. As seen in Table \ref{tab:before-after-hotels}, this caused 4,563 of the classes to be fall into one class.



\begin{table}[]
\caption{Number of images, branches (classes), and chains (super-classes) in the original and cleaned version of the dataset.}
\label{tab:before-after-hotels}
\small
\begin{tabular}{c|ccc}
\small
       & Images & Branches & Chains \\ \hline
Hotels-50K & 1,140,383    & 50,000         & 91           \\
\hotelsfifty  & 1,071,953    & 45,438         & 91          \\
\hotelsid  & 97,553    & 7,769   & 87         
\end{tabular}
\end{table}

\begin{table}[]
\vspace{-15pt}
\centering
\caption{\hotelsfifty: proposed splits with the number of branches (classes) and chains in each data split which determine four different settings explained in Figure \ref{fig:hotels50k-auc-barplot}.
}
\label{tab:hotels-splits}
\small
\begin{tabular}{c|ccc}
       & Images  & Branches & Chains \\ \hline
Train  & 294,275 & 17,281   & 81                                                                \\
Trainval & 348,461 & 18,918 & 84
                            \\ \hline
\firstval  & 22,121  & 4,941    & 79                                                                 \\
\firsttest & 51,294  & 11,532   & 81                                                                 \\ \hline
\secondval  & 16,095  & 740      & 71                                                                \\
\secondtest & 36,537  & 1,728    & 79                                                                \\ \hline
\thirdval  & 15,970  & 897      & 3                                                                  \\
\thirdtest & 35,693  & 1,365    & 7                                                                 \\ \hline
\fourthtest & 41,437  & 1,640    & -   \\ \hline
Total & 471,985 & 35,003 & 91
\end{tabular}\vspace{-10pt}
\end{table}
                                                               

\begin{table}[]
\vspace{-15pt}
\centering
\caption{\hotelsid: proposed splits for Hotel-ID number of branches (classes) and chains in each data split.}
\label{tab:hotelid-splits}
\small
\begin{tabular}{c|ccc}
       & Images  & Branches & Chains \\ \hline
Train  & 29,326 & 3,150   & 65     
                \\
Trainval  & 46,237 & 4,406   & 75     
                \\ \hline
\firstval  & 3,704  & 1,033    & 57     
\\
\firsttest & 9,013  & 3,698   & 74                                                                      \\ \hline
\secondval  & 6,612  & 617      & 56                                                                    \\
\secondtest & 11,110  & 881    & 61                                                                     \\ \hline
\thirdval  & 6,595  & 639      & 10                               
                    \\
\thirdtest & 10,973  & 737    & 12      
\\ \hline

\fourthtest & 20,220   & 1,745    & -    
\\ \hline
Total & 97,553   & 7,769     & 87    
                                                               
\end{tabular}\vspace{-10pt}
\end{table}

To reflect the characteristics needed for evaluating the models more realistically, we created one train split with 4 test sets, \firsttest, \secondtest, \thirdtest, and \fourthtest, divided based on the branches (classes) they contain. \firsttest consists of hotels branches that are also in the training set. \secondtest and \thirdtest consist of hotels branches that the training set does and does not contain their hotel \textit{chains}, respectively. And finally \fourthtest consists of hotel branches that the information about their hotel chain was not available and therefore, could be a mix of the first three test sets. The details of these splits can be seen in Table \ref{tab:hotels-splits} and \ref{tab:hotelid-splits}, which respectively correspond to the proposed \hotelsfifty and \hotelsid datasets. Note, that tests \firsttest, \secondtest, and \thirdtest  are increasingly closer to a real-world scenario depending on which classes have not been seen during training. As such, these tests may be increasing in mimicking a real-world scenario and also increasing in prediction difficulty. \fourthtest is unique in that the chain information is unknown; it may share similarities with tests \firsttest, \secondtest, or \thirdtest.
We also provide suggested validation sets ($\mathcal{D}^{val}_.$) for both datasets which have been been used for model selection and hyper-parameter tuning in this study. Note that since the validation sets are a subset of the ``Trainval'' set, images with ``unkonwn'' chain information cannot exist in the validation sets and hence, $\mathcal{D}^{val}_{??}$ does not exists for either datasets.

\section{Evaluation and Results}
\label{sec:res}



The goal of our experiments are twofold: 1) to evaluate the \hotelsfifty and \hotelsid datasets and their associated splits to highlight the increasing prediction difficulty in the test splits as we move to unseen (realistic) settings; 2) underscore the difference between the order of the methods among the splits.

\paragraph{Evaluation Setup} We fine-tune a ResNet50 \cite{he2016deep}, pre-trained on ImageNet \cite{deng2009imagenet}, for image retrieval with two types of loss functions. Namely, pair-based losses, where groups of datapoints are chosen and their embeddings are compared to eachother based on their labels, and proxy-based, where the model creates proxies for the classes and the datapoints are compared to the proxies rather than themselves. We use Triplet \cite{weinberger2006distance}, Circle \cite{sun2020circle}, Multi-Similarity \cite{wang2019multi}, and Supervised Contrastive \cite{khosla2020supervised} Losses for the pair-based losses, and ProxyNCA++ \cite{teh2020proxynca++} and Soft Triple Losses \cite{qian2019softtriple} for the proxy-based losses. Implementation details are discussed in Appendix \ref{app:baselines} and \ref{app:t-procedure}.

\paragraph{Metrics} To evaluate the models, we use two metrics: recall at one (\recallatone), and area under receiver operating characteristic curve (\randauroc). \recallatone is generally used for image retrieval tasks and measures how well a model can find similar images given a query image. \randauroc, on the other hand, measures how well the model can predict  how similar a pair of images are and whether they are from the same class. For both metrics, we use the cosine similarity function of the image embeddings to calculate the similarity of any two image.



\begin{figure}
    \centering\vspace{-5pt}
    \includegraphics[width=0.98\linewidth]{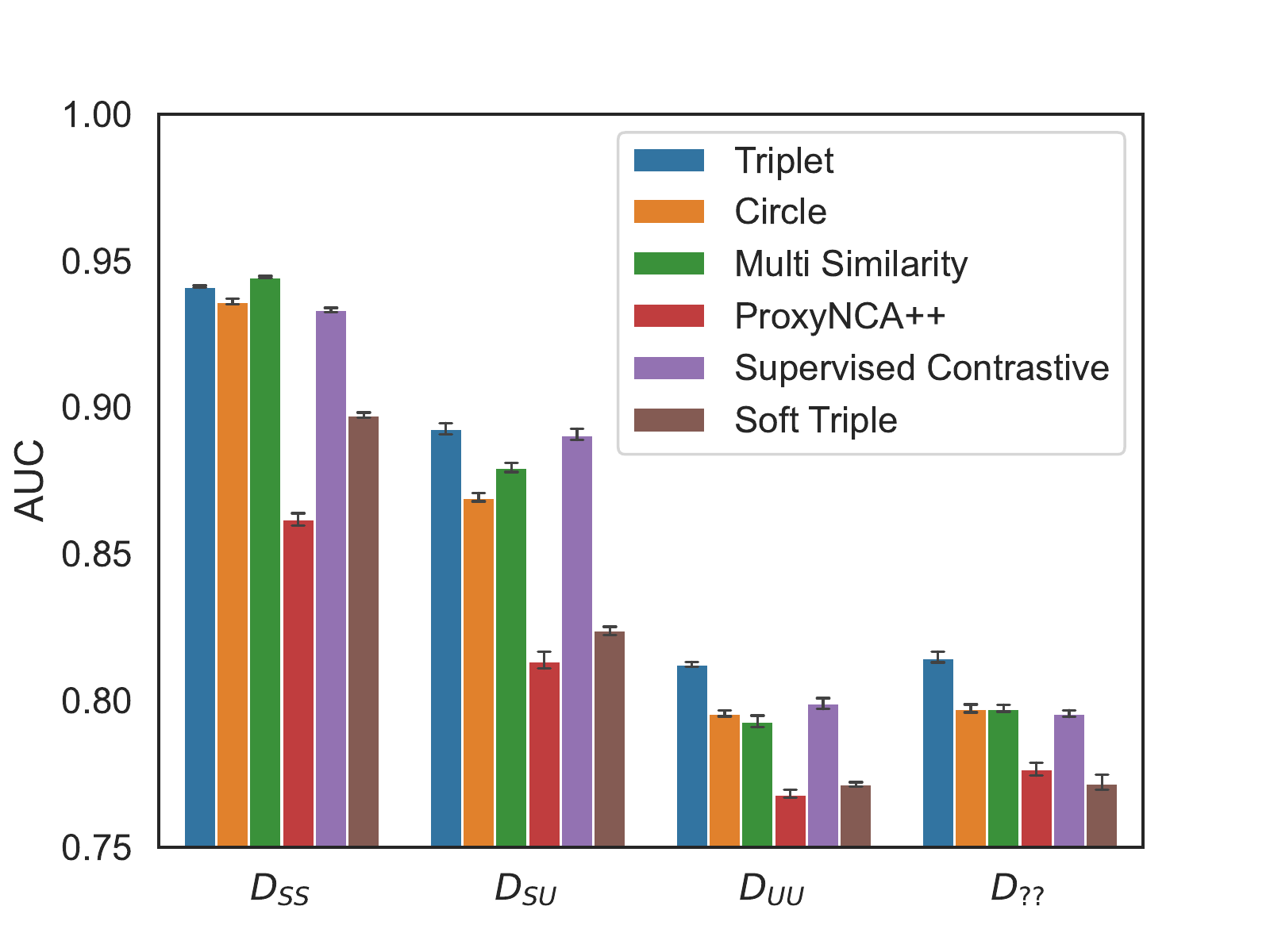}
    \vspace{-20pt}
    \caption{AUC in the proposed splits for \textbf{\hotelsid}, similar to \hotelsfifty. Here, the ranking of the models changes more significantly and the best performing model in the \firsttest is the forth method in \thirdtest.}
    \vspace{-15pt}
    \label{fig:hotelid-auc-barplot}
\end{figure}

\paragraph{\textbf{Results}}
In Figures \ref{fig:hotels50k-auc-barplot} and \ref{fig:hotelid-auc-barplot}, we observe that the \randauroc score of different methods consistently decrease moving from \firsttest to \secondtest to \thirdtest, which is as expected. For \fourthtest -- which is an unknown setting -- the change in prediction difficulty is also unknown and unclear, i.e. it seems to be more difficult in \hotelsfifty but less so in \hotelsid dataset. Also, we observe the ranking of best performing methods change in different evaluation settings for both metrics, showing the fact that different models generalize differently to more realistic situations. The full performance results can be seen in detail in Tables \ref{tab:hotels50-results} and \ref{tab:hotelid-results} in the appendix.

The increasing difficulty can also be seen for the image retrieval task between \secondtest and \thirdtest. However, the reason \firsttest is counter-intuitively more difficult than the rest, is because of how the split was constructed; the generally small class size and also the high number of classes causes the retrieval task to be more difficult. More information with regard to the datasplit generation can be found in Appendix \ref{app:generation}.

\section{Conclusions}
\label{sec:concl}

In this paper we propose new data splits of the Hotels-50K and Hotels-ID datasets, based on whether or not the data classes and super-classes were seen during training. The idea is to generate more realistic data splits with varying prediction difficulties, and be able to better evaluate the generalizability of models. 

Next, we do a comprehensive comparison among the performance of recent state-of-the-art image retrieval models on the revised datasets. We show that (i) the order of best-performing methods differs among different splits; and (ii) the overall performance of the models drop as the test splits become more difficult where the test sets include unseen classes, showing the relevance of our proposed splits to compare models in a more realistic and challenging scenario. 
\pagebreak
\bibliographystyle{icml2022}
\bibliography{main}

\clearpage
\newpage

\section*{Appendix}
\appendix






We provide additional material to support the main paper. We provide additional information on how we construct the new dataset splits and the class size distributions. Further, we elaborate on the architecture and training procedure and the hyper-parameters chosen for each dataset and model.

\section{Related Work}
\label{sec:background}


\subsection{Counter Human Trafficking Methods}

We categorize studies in computer science that are focused on human trafficking (HT), into three main groups: studies detecting HT by analysing individual ads, studies that link multiple ads for uncovering organized activity, and finally, studies that propose counter HT-specific tools, e.g. datasets. We elaborate on these categories in the following sections.

\paragraph{Methods based on individual ad analysis}
Escort ads often contain text and a set of images. One popular line of anti-HT studies focuses on the information in individual ads and aims to classify whether the ads are suspicious of HT or not. These studies mostly merely focus on the textual data \cite{wang2019sex, alvari2016non, dubrawski2015leveraging} and ignore all visual cues. For instance, in \cite{wang2019sex}, the authors analyze the textual data of ads in the Trafficking-10k \cite{tong2017combating} dataset to predict the likelihood of a an ad relating to HT.
In \cite{tong2017combating}, a multi-modal model is proposed that incorporates both the ad text and images for HT detection. Their language network consists of a long short-term memory (LSTM) network 
, followed by a fully connected layer to output a text representation. Their vision network consists of a VGG network followed by fully-connected layers to calculate the image representations. Finally, the two type of representations are combined with a third network to detect if an ad is related to HT or not. Although this study uses the images, they mention the vision module does \textit{not} add much to the performance of the model and that the images could potentially be used more efficiently.  

\paragraph{Methods based on linking ads}
Another line of work regarding anti-HT aims to extract features from each ad and link multiple ads together by finding shared features and detect organized activity in a \textit{group} of ads. Most of these studies only use the textual signals to link ads, such as shared phone numbers or bi-grams. For example, \cite{li2018detection} uses unsupervised learning on the textual data in ads to detect groups of ads with similar templates, and further, connect similar templates to identify the signature pattern of organized activities.

There have also been studies that use the hashcodes of the images in the ads, as well as the text, as indicators for linking ads \cite{rabbany2018active, nagpal2015entity}. Nevertheless, by using image hashcodes, only ads with the \textit{exact} same image will be matched, and minimal changes to the image, e.g. cropping or resizing it, will cause the hashcode to be drastically different from the original image hashcode, and hence, go unnoticed according to these algorithms. 

\paragraph{Dataset analysis}
Most supervised and unsupervised methods in DML on images use ubiquitous datasets for evaluating their methods, such as the CUB-200-2011 \cite{wah2011caltech} or the ImageNet \cite{deng2009imagenet} datasets. These datasets may be suitable for general applications, but nevertheless, they may not be suitable in applications such as analyzing escort ad images. This is because escort ads are often taken in hotel rooms, where a) classes are not well-defined; and b) the images tend to have more visual components. Hence, they do not provide the a vision model with sufficient information to make predictions.

Trafficking-10k \cite{tong2017combating} dataset, a benchmark dataset for anti-HT studies, which includes the text and images of more than 10,000 trafficking ads labeled with the likelihoods of them being associated with HT is one of the main datasets used in previous studies. Later, a study \cite{stylianou2017traffickcam} presented a fully realized crowdsourcing platform, TraffickCam, that is a database of hotel images and enabled law enforcement investigators to query hotel images. TraffickCam also allows public users to contribute to the database and submit up to four images of the hotel room at which they stay, along with the hotel's information. Interestingly, \cite{stylianou2019hotels} uses TraffickCam and travel websites to collect over a million images from 50,000 hotels from around the world, presenting the Hotels-50K dataset, in order to improve the task of hotel recognition. They collect their hotel images from 91 major hotel chains along with many smaller hotel chains. A subset of this dataset, Hotel-ID \cite{kamath20212021}, was also released, which only contains images from TraffickCam. This smaller dataset compared to the Hotels-50K, contains 7,769 hotels, and also 87 hotel chains.
However, despite being very rich in information, this dataset was originally created for hotel classification; hence, all classes in the evaluation section of it were previously seen by the model during training. Thus, using the standard training/evaluation splits does not reflect the real application scenario, where we can encounter unseen classes.



\begin{table*}[ht!]
\caption{Performance on \hotelsfifty Dataset.
For information regarding the splits, please refer to Table \ref{fig:hotels50k-auc-barplot}.
We mark the best performing score in \textbf{bold} and the second best performing is \underline{underlined}
}
\begin{center}
 \resizebox{0.96\linewidth}{!}{%
\begin{tabular}{clllllllll}
    \hline
    \multicolumn{1}{|c}{}    & \multicolumn{2}{c}{\firsttest }          &      \multicolumn{2}{c}{\secondtest}            &   \multicolumn{2}{c}{\thirdtest}             &       \multicolumn{2}{c|}{\fourthtest}      \\
    \hline
    \multicolumn{1}{|c|}{Metrics}  &  \recallatone   &   \multicolumn{1}{c|}{\randauroc} & \recallatone   &   \multicolumn{1}{c|}{\randauroc}  & \recallatone   &    \multicolumn{1}{c|}{\randauroc}  & \recallatone   &    \multicolumn{1}{c|}{\randauroc}  \\
    \hline    \hline
    \multicolumn{1}{|c|}{Triplet \cite{weinberger2006distance}}  &   10.84    &  \multicolumn{1}{l|}{95.05 $\pm$ 0.03}     &       31.34     &  \multicolumn{1}{l|}{ 93.65 $\pm$ 0.09}   &       28.04    &  \multicolumn{1}{l|}{87.90 $\pm$ 0.09}  &       18.38     &  \multicolumn{1}{l|}{80.90 $\pm$ 0.09} \\
    \multicolumn{1}{|c|}{Circle \cite{sun2020circle}}  &   26.14    &  \multicolumn{1}{l|}{\underline{95.39 $\pm$ 0.05}}      &       53.51     &  \multicolumn{1}{l|}{93.49 $\pm$ 0.06}   &       49.19    &  \multicolumn{1}{l|}{87.83 $\pm$ 0.06}  &       36.96   &  \multicolumn{1}{l|}{81.93 $\pm$ 0.08} \\
    \multicolumn{1}{|c|}{MultiSim \cite{wang2019multi}}  &   27.41   &  \multicolumn{1}{l|}{\textbf{96.13 $\pm$ 0.05}}    &       65.27   &   \multicolumn{1}{l|}{\textbf{94.25 $\pm$ 0.09}}   &       63.92   &   \multicolumn{1}{l|}{\textbf{88.99 $\pm$ 0.06}}  &       \underline{61.81}     &  \multicolumn{1}{l|}{\textbf{84.09 $\pm$ 0.10}} \\
    \multicolumn{1}{|c|}{ProxyNCA++ \cite{teh2020proxynca++}}  &   \textbf{42.58}   &  \multicolumn{1}{l|}{94.17 $\pm$ 0.07}   &       \textbf{72.40}   &   \multicolumn{1}{l|}{92.42 $\pm$ 0.05}   &       \textbf{67.85}   &  \multicolumn{1}{l|}{87.36 $\pm$ 0.07}  &       60.97   &   \multicolumn{1}{l|}{\underline{82.98 $\pm$ 0.10}} \\
    \multicolumn{1}{|c|}{SupCon \cite{khosla2020supervised}}  &   20.53   &  \multicolumn{1}{l|}{95.05 $\pm$ 0.04}   &       46.99   &  \multicolumn{1}{l|}{\underline{93.76 $\pm$ 0.09}}   &       39.45   &  \multicolumn{1}{l|}{\underline{88.54 $\pm$ 0.07}}  &       26.69   &  \multicolumn{1}{l|}{82.59 $\pm$ 0.11} \\
    \multicolumn{1}{|c|}{SoftTriple \cite{qian2019softtriple}}  &   \underline{36.54}   &  \multicolumn{1}{l|}{87.66 $\pm$ 0.07}   &       \underline{70.72}   &  \multicolumn{1}{l|}{86.73 $\pm$ 0.07}   &       \underline{67.70}   &  \multicolumn{1}{l|}{83.04 $\pm$ 0.09}  &       \textbf{62.87}   &  \multicolumn{1}{l|}{80.96 $\pm$ 0.12} \\
    \hline
\end{tabular}
}
\end{center}
\label{tab:hotels50-results}
\end{table*}

\begin{table*}[ht!]
\caption{Performance on \hotelsid Dataset. For information regarding the splits, please refer to Table \ref{fig:hotels50k-auc-barplot}.  We mark the best performing score in \textbf{bold} and the second best performing is \underline{underlined}.}
\begin{center}
 \resizebox{\linewidth}{!}{%
\begin{tabular}{clllllllll}
    \hline
    \multicolumn{1}{|c}{}    & \multicolumn{2}{c}{\firsttest}            &      \multicolumn{2}{c}{\secondtest}            &   \multicolumn{2}{c}{\thirdtest}             &       \multicolumn{2}{c|}{\fourthtest}      \\
    \hline
    \multicolumn{1}{|c|}{Metrics}  &  \recallatone   &  \multicolumn{1}{c|}{\randauroc} & \recallatone   &  \multicolumn{1}{c|}{\randauroc}  & \recallatone   &  \multicolumn{1}{c|}{\randauroc}  & \recallatone   &  \multicolumn{1}{c|}{\randauroc}  \\
    \hline
    \hline
    \multicolumn{1}{|c|}{Triplet \cite{weinberger2006distance}}  &   13.48 &  \multicolumn{1}{l|}{\underline{93.64 $\pm$ 0.13}} &       28.50     &  \multicolumn{1}{l|}{\underline{87.54 $\pm$ 0.14}}     &       23.43        &  \multicolumn{1}{l|}{\textbf{79.87 $\pm$ 0.20}}  &       15.38   &  \multicolumn{1}{l|}{\textbf{79.88 $\pm$ 0.12}} \\           
    \multicolumn{1}{|c|}{Circle \cite{sun2020circle}}           &   18.43             &  \multicolumn{1}{l|}{93.62 $\pm$ 0.11}             &       39.09                 &  \multicolumn{1}{l|}{86.93 $\pm$ 0.17}                 &       36.28     &  \multicolumn{1}{l|}{\underline{79.56 $\pm$ 0.11}}  &       26.27   &  \multicolumn{1}{l|}{79.74 $\pm$ 0.17} \\           
    \multicolumn{1}{|c|}{MultiSim \cite{wang2019multi}}         &   15.46    &  \multicolumn{1}{l|}{\textbf{94.27 $\pm$ 0.09}}    &       22.44        &  \multicolumn{1}{l|}{\textbf{87.71 $\pm$ 0.27}}        &       21.74                 &  \multicolumn{1}{l|}{79.45 $\pm$ 0.21}  &       11.51   &  \multicolumn{1}{l|}{\underline{79.76 $\pm$ 0.14}} \\     
    \multicolumn{1}{|c|}{ProxyNCA++ \cite{teh2020proxynca++}}   &   15.58                       &  \multicolumn{1}{l|}{86.21 $\pm$ 0.23}                       &       \textbf{52.94}                           &  \multicolumn{1}{l|}{81.35 $\pm$ 0.32}                           &       \textbf{48.15}                           &  \multicolumn{1}{l|}{76.81 $\pm$ 0.16}  &       \textbf{43.80}   &  \multicolumn{1}{l|}{77.67 $\pm$ 0.27} \\           
    
    \multicolumn{1}{|c|}{SupCon \cite{khosla2020supervised}}    &   15.69             &  \multicolumn{1}{l|}{92.85 $\pm$ 0.08}             &       23.74                 &  \multicolumn{1}{l|}{86.02 $\pm$ 0.28}                 &       18.26                 &  \multicolumn{1}{l|}{78.97 $\pm$ 0.13}  &       11.94   &  \multicolumn{1}{l|}{78.87 $\pm$ 0.18} \\                   
    \multicolumn{1}{|c|}{SoftTriple \cite{qian2019softtriple}}  &   \textbf{22.30}             &  \multicolumn{1}{l|}{89.56 $\pm$ 0.11}             &       \underline{47.99}                 &  \multicolumn{1}{l|}{81.08 $\pm$ 0.21}                 &       43.07                 &  \multicolumn{1}{l|}{76.31 $\pm$ 0.13}  &       \underline{35.72}   &  \multicolumn{1}{l|}{76.03 $\pm$ 0.17} \\                   
    \hline
\end{tabular}
}
\end{center}
\label{tab:hotelid-results}
\end{table*}

\subsection{Deep Metric Learning}

Deep metric learning (DML) on images was first introduced in \cite{bromley1993signature}, which proposed the Contrastive loss function for signature verification. Building on the Contrastive loss, the Triplet loss was introduced in \cite{weinberger2006distance} where, given a positive and negative pair with the same anchor datapoint, aims for the positive pair to be closer to each other than the negative pair by some fixed margin. The advantage of the Triplet loss over the Contrastive loss is that it optimizes the relative distance of the datapoints, rather than optimizing the distance of every two datapoints. 

Some of the main places of improvement for the triplet loss, are the method the triplets are chosen and also how the loss is calculated for each triplet. Since choosing every possible triplet is infeasible for large datasets, it is common to choose a fixed number of triplets \textit{randomly} to mitigate this issue. However, this method may not yield the most informative triplets for updating the model weights and studies tend to use more elaborate strategies for finding ``hard'' or ``semi-hard'' negatives for every anchor-positive pair \cite{schroff2015facenet}. Furthermore, inspired by the contrastive loss functions used in self-supervised learning studies \cite{wu2018unsupervised, chen2020simple}, \cite{khosla2020supervised} presents the ``\textit{supervised} contrastive loss'' where the loss function contrasts all the positive samples in a mini-batch with all the negatives in it. The authors prove that this loss is a generalized version of the Triplet loss because instead of selecting one positive and one negative, an arbitrary number of positives and negatives can be used per anchor to enhance the model's performance, addressing the drawbacks mentioned for the triplet loss. 
Another study tackling these issues is \cite{wang2019multi}, in which the authors define three types of pair-based similarities: self-similarity, positive-similarity, and negative-similarity. 
They show none of the previous pair-based losses, e.g., contrastive, and triplet, encompasses all three similarity types and introduce the Multi-Similarity Loss which utilizes all three to weigh samples and calculate the loss among them.


To mitigate the sampling challenges posed by triplet-based methods, the proxy-based losses were proposed. A proxy is defined as a representative of a subset of the dataset and learned as a part of the model while training. In addition, the distance of datapoints are calculated with respect to the proxies, rather than the other datapoints directly, causing the cost of calculating all possible distances to drastically decrease. Examples of these types of losses are \cite{kim2020proxy, teh2020proxynca++}, where one proxy is assigned to each class in the dataset, which aims to maximize the odds of a datapoint belonging the correct proxy. In other words, the distance between any datapoint and its respective class's proxy should be less than all other proxies. 

Another workaround for the challenges of pair- and triplet-based losses is using classification losses such as cross entropy (CE) for training the model on a classification task, and using the learnt model for producing image representations. An example of these studies is \cite{noh2017large}, where the model is trained on a classification task, and after being trained, it is fine-tuned with an attention module. For image retrieval during inference, the embeddings created by the penultimate layer are used as the feature descriptors of the images. \cite{jun2019combination} is another example that aims to perform image retrieval by initially using a classification loss for fine-tuning and then train their model with a Triplet loss. They mention the classification loss causes the inter-class distances to maximize, which helps the convergence of the model while using the Triplet loss. Inspired by the classification loss, \cite{qian2019softtriple} proposes to assign multiple centers for each class, as opposed to one, and calculate the classification loss based on any samples distance to all class centers. They further show that this loss is a generalized version of the Triplet Loss.

Although image retrieval has been extensively studied, these methods may not yield the most efficient algorithms for performing image linking. The main reason is that since the labels are defined on images, rather than image \textit{pairs}, when learning datasets with high intra-class variance, the model is forced to map relatively different images with the same label to a single class, which may cause the model to memorize features rather than learn them. However, since image linking defines the labels on image pairs, connected images definitely share a commonalty, causing the model to learn the shared feature between them.




\section{Datasets Related}
\label{app:generation}
\subsection{Splitting method}
Recall from Section \ref{sec:hotels}, that our objective is to create more realistic evaluation splits compared to the original datasets, in that some hotel chains (i.e., ``super-classes'') or branches (i.e., ``classes'') that appear during test should have not to been seen during training. Our overall strategy is to first separate the evaluation splits based on their difficulty requirements; then whatever is left from the datasets will be assigned to their respective training set. 

We utilize the information from both the super-classes and classes. In each dataset, we define $\mathcal{B}$ and $\mathcal{C}$ as a set of all hotel branches and chains, respectively. We first separate all hotels branches with unknown \textit{chains} in $\mathcal{C}$, and generate $\mathcal{C}_{known}$ and $\mathcal{C}_{unknown}$ as the set of all known and unknown chain labels, respectively. Based on this initial separation, we form $\mathcal{B}_{known} = \{b \mid \text{chain}(b) \in \mathcal{C}_{known}\}$ and $\mathcal{B}_{unknown} = \{b \mid text{chain}(b) \in \mathcal{C}_{unknown}\}$, the set of hotel \textit{branches} with a known and unknown hotel chain, respectively. 

Recall that \fourthtest is meant to contain images from all unknown chains. Hence, we assign all images with their hotel branch in $\mathcal{B}_{unknown}$ to \fourthtest. 

\thirdtest, \secondtest, and \firsttest are subsets of all images associated with $\mathcal{C}_{known}$ based on whether or not the chains or branches are represented in the train split. \thirdtest contains images from unseen hotel chains and branches. Hence, we reserve a set of hotel chains $\mathcal{C}_{\thirdtest} \subset \mathcal{C}_{known}$ for \thirdtest, and assign all images with branches in $\mathcal{B}_{\thirdtest} = \{b \mid \text{chain}(b) \in \mathcal{C}_{\thirdtest}\}$ to \thirdtest. 

We construct \secondtest, \firsttest, and the train split from images with branches assigned to a chain in $\mathcal{C}_{rest} = \mathcal{C}_{known} - \mathcal{C}_{\thirdtest}$ and branches in $\mathcal{B}_{rest} = \{b \mid b \in \mathcal{C}_{rest}\}$. (This ensures that \thirdtest contains images from hotel chains that are not seen during training.) For these splits, we enforce (i) $\mathcal{C}_{train} = \mathcal{C}_{rest}$, (ii) $\mathcal{C}_{\secondtest} \subseteq \mathcal{C}_{train}$ and (iii) $\mathcal{C}_{\firsttest} \subseteq \mathcal{C}_{train}$ to remain true in the subsequent steps. This ensures that the chains in \firsttest and \secondtest have been seen during training.

We create \secondtest by assigning \textit{all} images from a random sample of hotel branches $\mathcal{B}_{\secondtest}$ out of $\mathcal{B}_{rest}$. Doing so guarantees that although the hotel chains in this split is a subset of the chains in the train, but the branches are exclusive.

Finally, we divide the unassigned images into the train and \firsttest. \firsttest contains a sample of branches and chains which are seen during training. Hence, $\mathcal{B}_{\firsttest} \subseteq \mathcal{B}_{train}$. Importantly, we first filter out branches with images fewer than a threshold, $t_1$, to only be used for training. This ensures that each hotel branch in \firsttest also has a \textit{sufficient} number of images associated with that branch in the train set. 

We are left with branches $B = \{b_1, b_2, ..., b_n\}$, each containing $N_i$ images where $N_i \geq t_1$ and $i \in \{1, 2,... n\}$. We define a second threshold, $t_2$ as the \textit{minimum} number of images each branch in \firsttest should contain. Next, for each branch $b_i$, we sample a random number, $k_i \in \{t_2, t_2 + 1, ..., \frac{N_i}{5}\}$. (5 is chosen arbitrarily, but was empirically found to maintain a reasonable test-to-train ratio across the four test sets). Finally, for each branch, we assign $k_i$ images to \firsttest. All remaining images are assigned to the train set.

\subsection{Statistics}
\begin{figure}
    \centering
    \includegraphics[width=\linewidth]{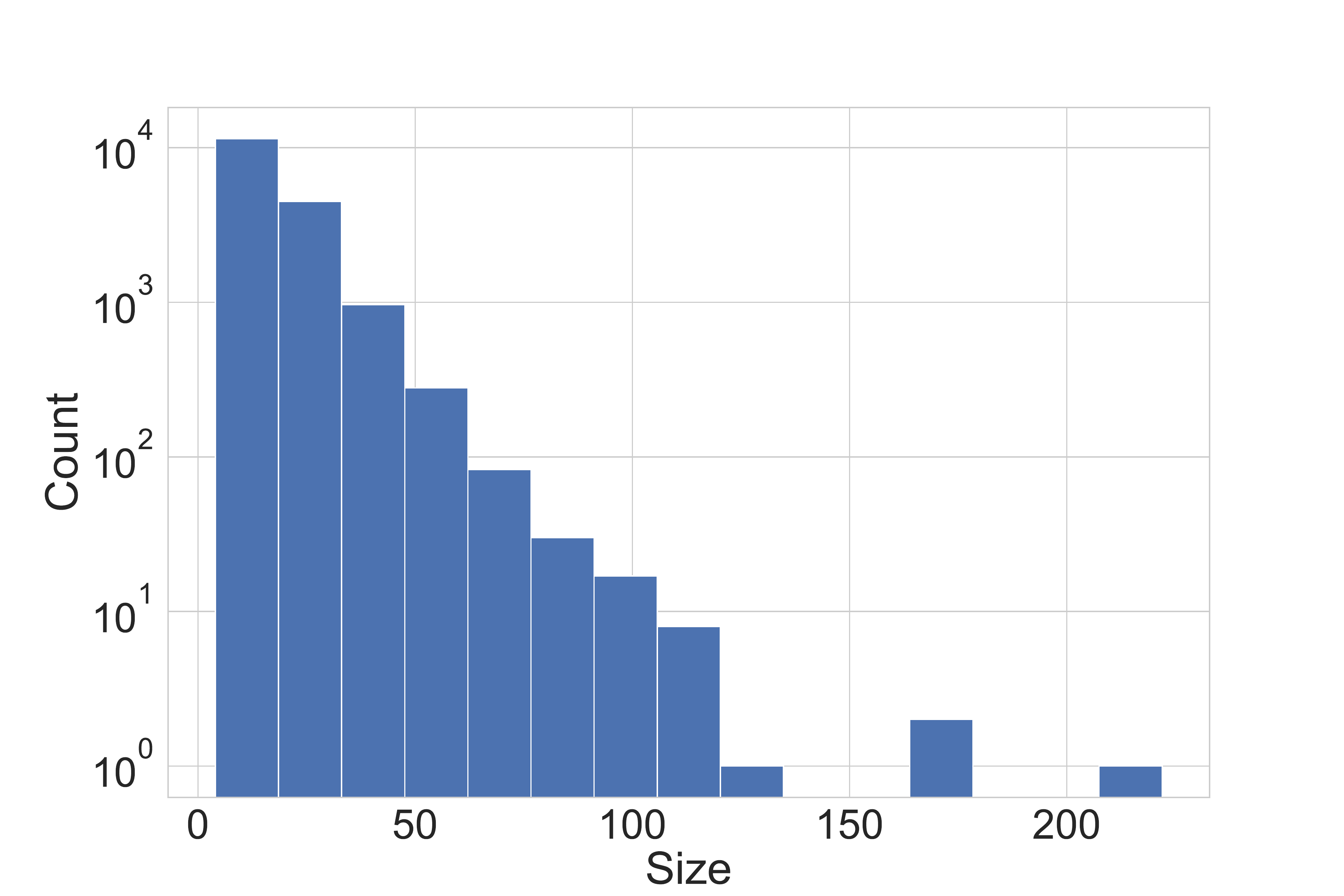}
    \caption{Class sizes for \hotelsfifty Trainval split.}
    \label{fig:hotels50-dist-train}
\end{figure}

\begin{figure}
    \centering
    \includegraphics[width=\linewidth]{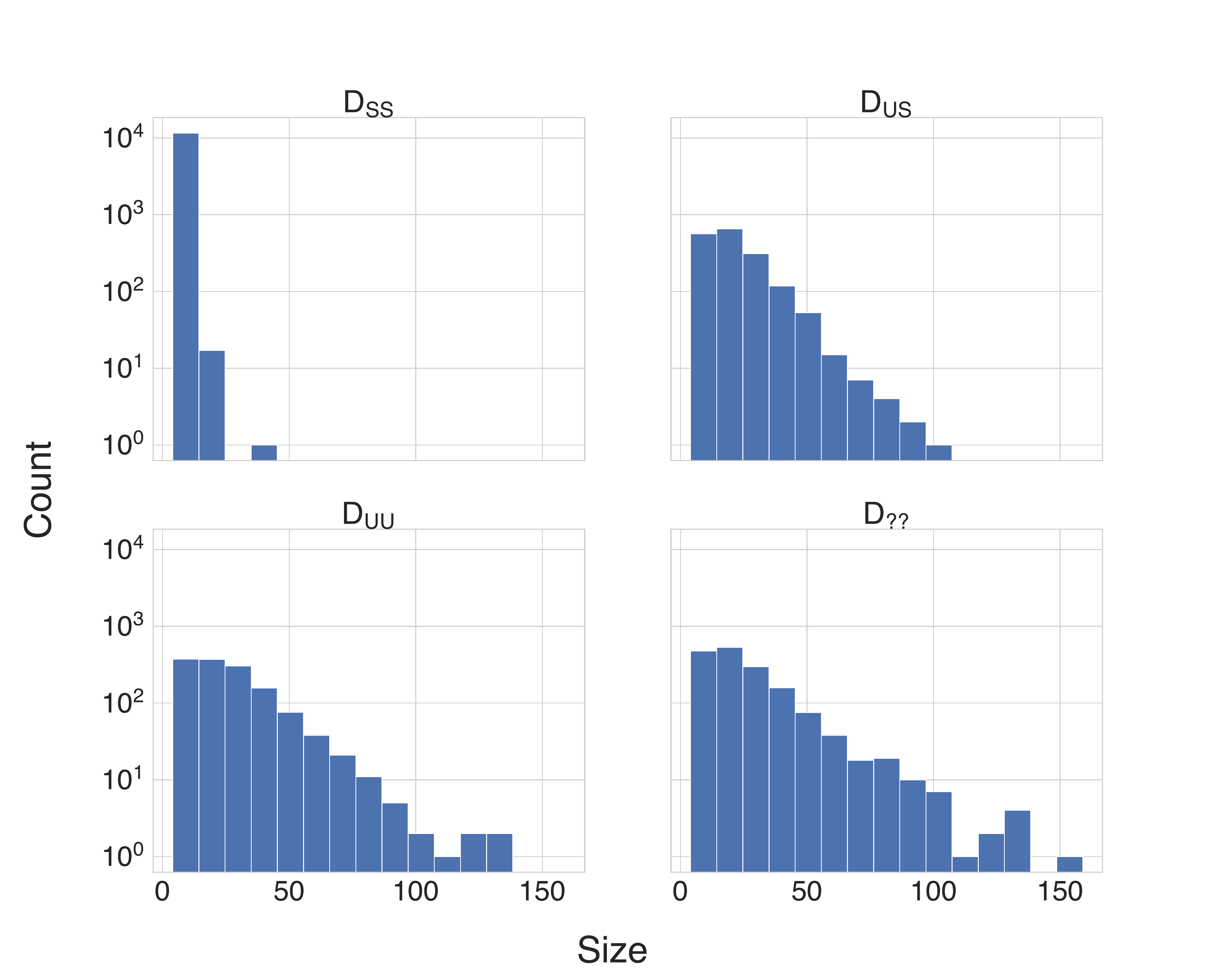}
    \caption{Class sizes for \hotelsfifty Evaluation splits.}
    \label{fig:hotels50-dist-eval}
\end{figure}

\begin{figure}
    \centering
    \includegraphics[width=\linewidth]{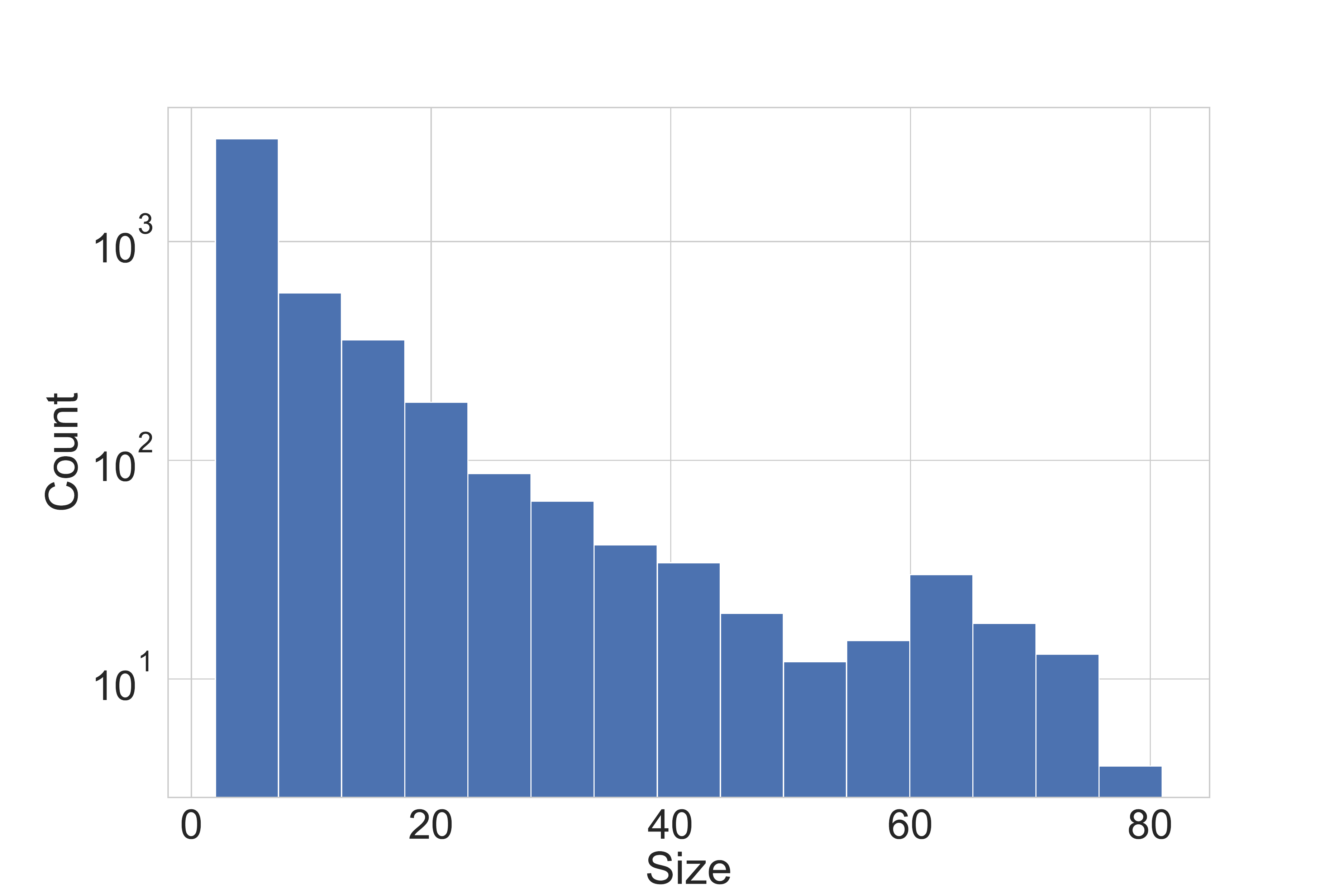}
    \caption{Class sizes for \hotelsid Trainval split.}
    \label{fig:hotelid-dist-train}
\end{figure}

\begin{figure}
    \centering
    \includegraphics[width=\linewidth]{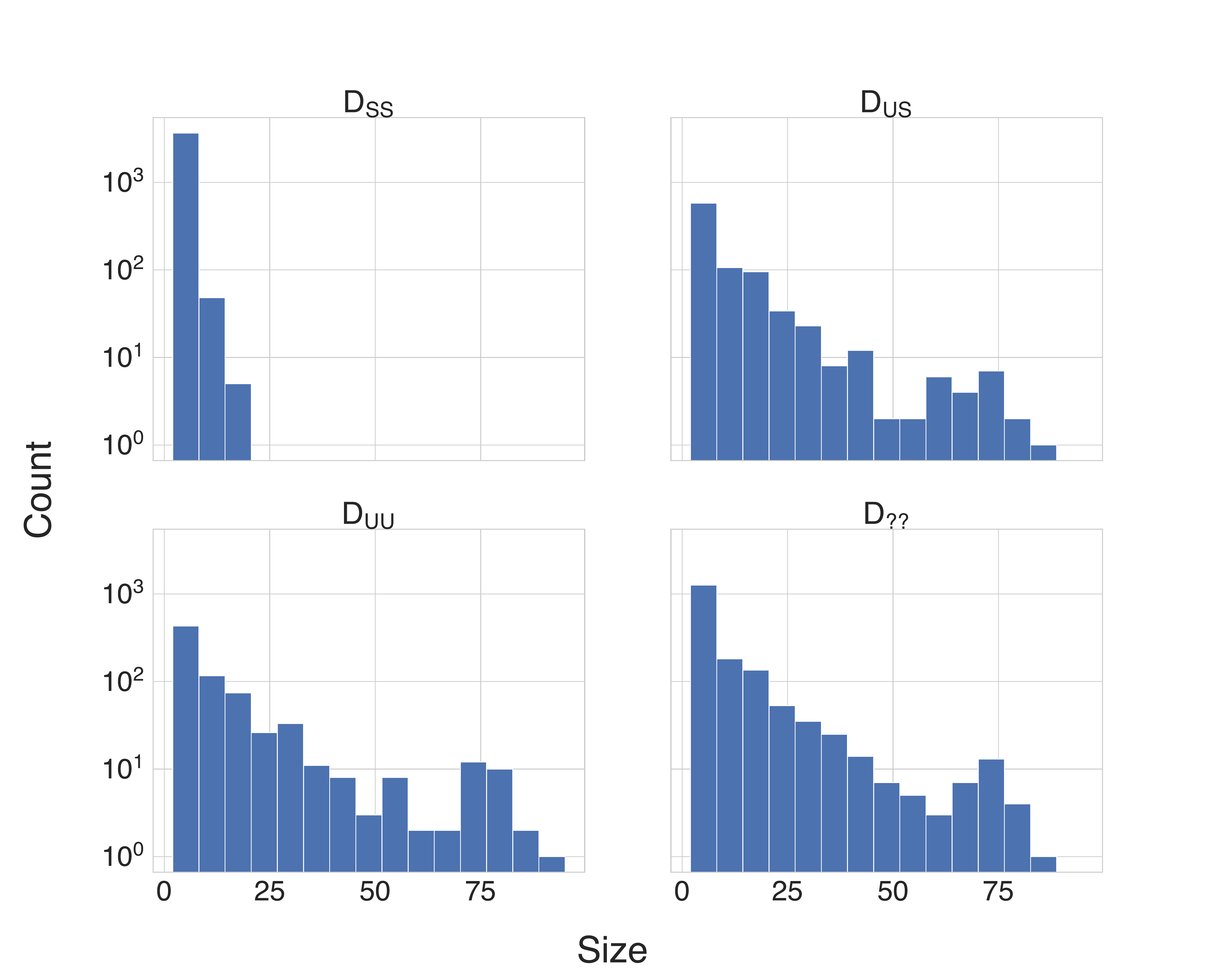}
    \caption{Class sizes for \hotelsid Evaluation splits.}
    \label{fig:hotelid-dist-eval}
\end{figure}

We plot the distributions of the evaluation and training set of \hotelsfifty and \hotelsid in Figures \ref{fig:hotels50-dist-train}, \ref{fig:hotels50-dist-eval}, \ref{fig:hotelid-dist-train}, and \ref{fig:hotelid-dist-eval}. As it can be seen, the dataset sizes have a long-tail distribution and therefore, for the sake of clarity, we use a log-scale y-axis in both datasets. Since \secondtest, \thirdtest, and \fourthtest contain all images from their respective hotel branches, their sizes follow have a longer tail. We see that due to the method \firsttest was created, there are a large number of classes with very few images in them. This is the main reason the image retrieval metric is more difficult on this split that \secondtest, \thirdtest, and \fourthtest.

\section{Image Retrieval Baselines}
\label{app:baselines}

We employ a model ($\Phi$) consisting of a pre-trained Resnet50 network as the backbone followed by a fully connected layer for projection. We use the following commonplace image retrieval loss functions to train the model:



\begin{itemize}
    \item \textbf{Triplet} \cite{weinberger2006distance}: Given $A$ and $P$ from the same class and $N$ from a different one, pushes the embeddings $A$, $P$, and $N$ such that the $S(A, P)$ similarity is larger than $S(A, N)$ according to a predefined similarity function $S(., .)$. Assuming $S(., .)$ is cosine similarity, we define $s_{ij}^p = \s{x^p_j}{x_i}$ and $s_{ik}^n = \s{x^n_k}{x_i}$, where for every example $x_i$, $x^p_j$ is the $j^{th}$ example of the positive set $\mathcal{P}_i$ and $x^n_k$ is the $k^{th}$ example of the negative set $\mathcal{N}_i$.
    
    
    \item \textbf{Circle} \cite{sun2020circle}: In addition to sampling triplets similar to Triplet loss and calculating the distances, assigns a different weight to the $D(A, P)$ and $D(A, N)$.


    
    
    
    
    
    \item \textbf{ProxyNCA++} \cite{teh2020proxynca++}: Learns a proxy embedding per class, and given a sample from class $C$ and its distances to all proxies ($A$), and maximizes the probability of the sample belonging to class $C$. Given an example $x_i$, assuming that $f(x_i)$ is its proxy and $d(., .)$ is the Euclidean distance, the loss is calculated as:
    
    
    \item \textbf{SoftTriple} \cite{qian2019softtriple}: Assumes multiple centers for each class and for each sample, applies a classification loss to with respect to its closest class center. Embeddings of images and centers have the unit length in the experiments. Given a sample $x_i$, the similarity to class $c$ with $J$ centers is computed as
    
    
    
    \item \textbf{Supervised Contrastive} \cite{khosla2020supervised}: Given a batch of data, initially creates an augmented batch, namely the ``multiviewed batch'', by generating two augmentations of each sample and pushes all samples from the same class together. For $N$ randomly sampled samples, first the multiviewed batch is created, where $i \in I = \{1,2,...,2N\}$ is the index of arbitrary augmented samples and $A_i = I - {i}$. In this method, the image embeddings (e.g. $x_i$) are normalized onto the unit sphere.
    
    
    \item \textbf{Multi-Similarity}\cite{wang2019multi}: Samples and calculates the loss for any batch according to the three similarity types defined in their study.
    
    
\end{itemize}

\section{Training Details}
\label{app:t-procedure}

\subsection{Hardware and Architecture}

Where applicable, we train the network by passing it batches of size $|B| = m \times k$, where $m$ is the number of distinct classes and $k$ is the number of instances per class.\footnote{When using the ProxyNCA++ loss function we use the official provided GitHub repository which slightly differs in how the the batches are constructed.}

To implement and evaluate the baseline models, we follow the architecture used in \cite{teh2020proxynca++}; namely a ResNet50 \cite{he2016deep} followed by a Layer Normalization \cite{ba2016layer} and a an additional fully connected layer of size 512 for projection. Besides the ``BCE'' and ``BCE Hard'' losses, which we implement ourselves, we use the loss functions implemented by the ``PyTorch Metric Learning'' library \cite{musgrave2020pytorch} for the Triple, SoftTriple, MultiSim, Supervised Contrastive Loss and finally the code provided by \cite{teh2020proxynca++} on their provided GitHub repository for the ProxyNCA++ baseline. For implementing fast pairwise distance calculation for the sake of evaluation, with use the Faiss \cite{johnson2019billion} library. Finally, to train our models, we use a single NVIDIA GeForce RTX 2080 Ti for the experiments related to \hotelsfifty, and an NVIDIA Tesla V100-SXM2-32GB for the \hotelsid dataset.

\subsection{Hyper-Parameter Tuning}

All models were trained for 60 epochs and optimized while monitoring the \recallatone metric. We use the suggested validation sets (\firstval, \secondval, and \thirdval) for hyper-parameter tuning the models.
\begin{table}[ht]
\caption{Final hyper-parameters chosen for each dataset. This the columns represent the learning rate (LR), batch size (BS), and other hyper-parameters specific to the method, namely proxy learning rate (plr), centers per class (clc), temperature (temp), and number of classes per batch (nc). }
\begin{center}
 \resizebox{0.99\linewidth}{!}{%
\begin{tabular}{cllllll}
    \hline
    \multicolumn{1}{|c}{}    & \multicolumn{3}{c}{\hotelsfifty}            &      \multicolumn{3}{c|}{\hotelsid}      \\
    \hline
    \multicolumn{1}{|c|}{Hyper Parameters}  &  LR   &   BS   &  \multicolumn{1}{c|}{Other} & LR   &   BS   &  \multicolumn{1}{c|}{Other}  \\
    \hline
    \hline
    \multicolumn{1}{|c|}{Triplet \cite{weinberger2006distance}}  &   0.00001   &   80   &  \multicolumn{1}{l|}{-}     &       0.00003   &   80   &  \multicolumn{1}{l|}{-}    \\
    \multicolumn{1}{|c|}{Circle \cite{sun2020circle}}  &   0.0001   &   80   &  \multicolumn{1}{l|}{-}     &       0.0001   &   80   &  \multicolumn{1}{l|}{-}    \\
    \multicolumn{1}{|c|}{MultiSim \cite{wang2019multi}}    &   0.00003   &   80   &  \multicolumn{1}{l|}{-}     &       0.0001   &   256   &  \multicolumn{1}{l|}{-}    \\ 
    \multicolumn{1}{|c|}{ProxyNCA++ \cite{teh2020proxynca++}}  &   0.024   &   96   &  \multicolumn{1}{l|}{\begin{tabular}[c]{@{}l@{}}plr: 240\\ nc: 32\end{tabular}}     &       0.024   &   192   &  \multicolumn{1}{l|}{\begin{tabular}[c]{@{}l@{}} plr: 240\\ nc: 64\end{tabular}}  \\
    \multicolumn{1}{|c|}{SupCon \cite{khosla2020supervised}}   &    0.00001   &   80   &  \multicolumn{1}{l|}{temp: 0.05}     &       0.00003   &   256   &  \multicolumn{1}{l|}{temp: 0.05}    \\
    \multicolumn{1}{|c|}{SoftTriple \cite{qian2019softtriple}}    &   0.00003   &   64   &  \multicolumn{1}{l|}{cpc: 5}     &       0.00003   &   64   &  \multicolumn{1}{l|}{cpc: 20}    \\
    \multicolumn{1}{|c|}{BCE}       &   0.0003   &   80   &  \multicolumn{1}{l|}{-}     &        0.0001   &   128   &  \multicolumn{1}{l|}{-}    \\
    \multicolumn{1}{|c|}{Hard BCE}    &    0.00001   &   80   &  \multicolumn{1}{l|}{-}     &        0.0001   &   256   &  \multicolumn{1}{l|}{-}    \\
    \hline
\end{tabular}
}
\end{center}
\label{tab:hyperparams}
\end{table}

For the methods with loss-related hyper-parameters, we use the suggested hyper parameters they mention in their respective paper - unless explicitly mentioned in this section. For training all models with both datasets, we search all learning rates in $\{3 \times 10^{-2}, 10^{-2}, 3 \times 10^{-3}, 10^{-3}, 3 \times 10^{-4}, 10^{-4}, 3 \times 10^{-5}, 10^{-5}, 3 \times 10^{-6}, 10^{-6}, 3 \times 10^{-7}\}$. For finding the optimal batch size for each dataset, we search batch sizes in $\{64, 80\}$ for \hotelsfifty and batch sizes of size $\{64, 80, 128, 256\}$ for \hotelsid. In the case of the Triplet Loss, we search margins in $\{0.05, 0.1, 0.15, 0.2\}$ and in case of the SupCon, we search temperatures in $\{0.05, 0.1, 0.2\}$. With the SoftTriple loss, since it assigns learns \textit{multiple} centers of the save embedding size for each class, due to the large number of classes in \hotelsfifty, we searched $\{5, 10\}$ for \hotelsfifty, and $\{5, 10, 15, 20\}$ for \hotelsid. Finally, with ProxyNCA++ \cite{teh2020proxynca++}, besides the mentioned batch sizes for both datasets, we also try batch size 96 for \hotelsfifty, and batch size 192 for \hotelsid, and learning rates suggested by their paper for both datasets. The details of the hyper-parameters used for each method can be seen in Table \ref{tab:hyperparams} and the results can be seen in Tables \ref{tab:hotels50-results} and \ref{tab:hotelid-results}.

\end{document}